\documentclass[times, review, 10pt]{elsarticle}

\usepackage{amssymb}
\usepackage{amsmath}
\usepackage[table]{xcolor}%
\usepackage{multirow}
\usepackage{url}

\journal{Nuclear Physics B}

\begin{document}

\begin{frontmatter}



\title{Dual-level Modality Debiasing Learning for Unsupervised Visible-Infrared Person Re-Identification\tnoteref{grant}}
\tnotetext[grant]{This work is supported by the National Natural Science Foundation of China (Grant No. 62272430).}

\author[label1]{Jiaze~Li\fnref{1}} 
\ead{jz_li@mail.ustc.edu.cn}


\affiliation[label1]{organization={University of Science and Technology of China}, 
                city={Hefei},
                postcode={230026}, 
                country={China}}

\author[label2]{Yan~Lu\fnref{1}} 
\ead{luyan@pjlab.org.cn}

\fntext[1]{These authors contributed equally to this work.}


\affiliation[label2]{organization={Shanghai Artificial Intelligence Laboratory}, 
                city={Shanghai},
                postcode={200233}, 
                country={China}}

\author[label1]{Bin~Liu\corref{cor1}} 
\ead{flowice@ustc.edu.cn}
\cortext[cor1]{Corresponding author}

\author[label1]{Guojun~Yin} 
\ead{gjyin@mail.ustc.edu.cn}

\author[label3]{Mang~Ye} 
\ead{yemang@whu.edu.cn}
\affiliation[label3]{organization={the School of Computer Science, Wuhan University}, 
                city={Wuhan},
                postcode={430072}, 
                country={China}}


\begin{abstract}
Two-stage learning pipeline has achieved promising results in unsupervised visible-infrared person re-identification (USL-VI-ReID).  It first performs single-modality learning and then operates cross-modality learning to tackle the modality discrepancy.
Although promising, this pipeline inevitably introduces modality bias: modality-specific cues learned in the single-modality training naturally propagate into the following cross-modality learning, impairing identity discrimination and generalization.
To address this issue, we propose a Dual-level Modality Debiasing Learning (DMDL) framework that implements debiasing at both the model and optimization levels. 
At the model level, we propose a Causality-inspired Adjustment Intervention (CAI) module that replaces likelihood-based modeling with causal modeling, preventing modality-induced spurious patterns from being introduced, leading to a low-biased model.
At the optimization level, a Collaborative Bias-free Training (CBT) strategy is introduced to interrupt the propagation of modality bias across data, labels, and features by integrating modality-specific augmentation, label refinement, and feature alignment.
Extensive experiments on benchmark datasets demonstrate that DMDL could enable modality-invariant feature learning and a more generalized model.
The code is available at \url{https://github.com/priester3/DMDL}.
\end{abstract}




\begin{keyword}
Visible-infrared person re-identification \sep Unsupervised learning \sep Causal intervention \sep Modality-invariant feature
\end{keyword}

\end{frontmatter}



\section{Introduction}
\label{sec:intro}

Visible-infrared person re-identification (VI-ReID) focuses on the identification and matching of individuals across distinct modalities, visible and infrared.
Remarkable progress has been made in this field, as evidenced by the success of existing works \cite{ye2021channel,ren2024implicit}. 
However, the collection of extensive cross-modality annotations is a costly and time-consuming process, which poses limitations on its broader application. As a solution, Unsupervised Visible-infrared Person Re-identification (USL-VI-ReID) \cite{yang2022augmented,wu2023unsupervised,cheng2023efficient} has emerged to facilitate VI-ReID without the reliance on human identity labels.

\begin{figure}[t]
    \centering
    \includegraphics[width=1.0\linewidth]{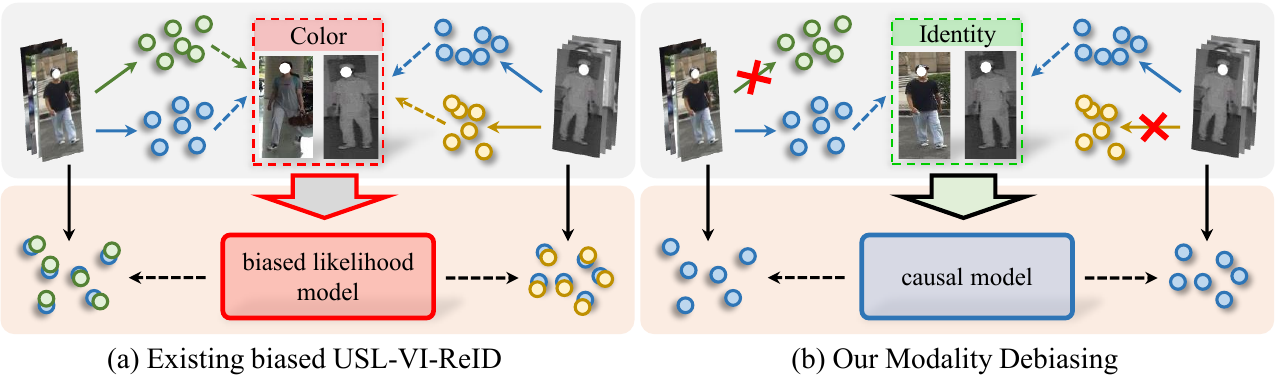}
    \caption{
    Existing USL-VI-ReID methods suffer from modality bias, leading to modality-related features. In contrast, our approach achieves modality-invariant feature learning through causal modeling and unbiased optimization. Green, yellow, and blue circles represent visible-specific, infrared-specific, and modality-shared information, respectively.
    }
    \label{fig:1}
\end{figure}

The main challenge in the USL-VI-ReID is the modality discrepancy, which limits the direct application of standard unsupervised learning of traditional unsupervised ReID. Therefore, the mainstream methods for USL-VI-ReID typically follow a two-stage learning pipeline \cite{wu2023unsupervised,cheng2023efficient,shi2024learning,teng2025relieving}:
1) In the first stage, the model is trained by operating unsupervised learning techniques \cite{dai2022cluster} on each modality separately to have the single-modality discriminative ability. 
2) In the second cross-modality unsupervised process, the model alternately establishes relationships across modalities and fits these relationships to achieve cross-modality discrimination capabilities.
Although promising, it also suffers from a modality bias issue that restricts the overall results. 
The first single-modality learning process naturally captures modality-specific cues from visible/infrared data, resulting in a biased model.
Initializing the second stage with this model inevitably introduces modality bias into the cross-modality learning, leading to biased cross-modality relationships, e.g., similar clothing color cues may result in incorrect matches across modalities, as illustrated in Fig.~\ref{fig:1} (a).
Since cross-modality relationships (i.e., pseudo labels) are the model-fitting target in the second stage, the biased knowledge (i.e., modality-specific cues) is gradually enhanced in the learned patterns, leading to modality-related features. In summary, modality bias originating from data propagates into labels and features throughout the learning pipeline, leading the model to rely on modality-specific cues for identification and thereby significantly limiting its generalization.

To address the aforementioned modality bias issue, we propose a Dual-level Modality Debiasing Learning (DMDL) framework.
DMDL performs modality debiasing at both the model and optimization levels, where the former prevents the model from learning modality bias in incorrect cross-modality relationships, and the latter aims to disrupt the propagation of biased knowledge from data to labels and features directly.
To this end, a Causality-inspired Adjustment Intervention (CAI) module and a Collaborative Bias-free Training (CBT) strategy are proposed. 
Specifically, CAI facilitates causal intervention under cross-modality unsupervised learning with backdoor adjustment, making the model only capture the causal patterns.
Compared with the traditional likelihood method, the causal modeling in CAI is theoretically unaffected by modality bias, thereby achieving a low-biased model. 
To further prevent biased knowledge from deepening during optimization, we propose the CBT strategy, integrating data augmentation, label refinement, and feature alignment. 
CBT first introduces a pseudo-modality augmentation scheme to modify modality-specific cues in images. Based on the augmented images, a cross-modality label smoothing scheme and a feature alignment loss are proposed to refine the biased relationships and learn shared knowledge across pseudo-modalities, respectively.
By jointly leveraging these components, CBT explicitly interrupts the propagation of modality bias across data, labels, and features.
Ultimately, the overall DMDL keeps an effective modality debiasing implementation, achieving modality-invariant feature learning, as Fig.~\ref{fig:1} (b) shows.

Our main contributions are summarized as follows: 

\begin{enumerate}[(1)]

\item We investigate the modality bias issue for existing USL-VI-ReID methods and propose a Dual-level Modality Debiasing Learning (DMDL) framework performed at both the model and optimization levels to learn modality-invariant feature representations.

\item We propose a Causality-inspired Adjustment Intervention (CAI) module at the model level to effectively model the causal patterns, constructing a low-biased model.

\item We propose a Collaborative Bias-free Training (CBT) strategy at the optimization level, combining label refinement and feature alignment with modality-specific data augmentation to prevent fitting biased knowledge.

\item Extensive experiments conducted on standard visible-infrared ReID benchmarks demonstrate the effectiveness and superiority of our method. 

\end{enumerate}

\section{Related Work}

\subsection{Unsupervised Visible-Infrared Person ReID}
Traditionally, visible–infrared ReID and unsupervised ReID were studied as two largely independent tasks. For both image-level~\cite{li2025shape} and video-level~\cite{li2025video,leng2025dual} VI-ReID, the core objective is to construct a cross-modality identity-discriminative space that is consistent across visible and infrared domains. In contrast, unsupervised ReID~\cite{li2022mutual,li2023logical} typically focuses on exploiting multi-view information or local feature interactions to generate reliable pseudo labels, thereby enabling the learning of discriminative representations without manual annotations.

By integrating these two paradigms, USL-VI-ReID naturally emerges as a promising research direction without requiring any human annotations.
Most existing approaches adopted a two-stage pipeline to mitigate the significant modality discrepancy, and most of them aimed at exploring reliable cross-modality correspondences.
For instance, PGM~\cite{wu2023unsupervised} and MBCCM~\cite{cheng2023efficient} utilized graph matching to establish reliable relationships across modalities globally, while DOTLA~\cite{cheng2023unsupervised} leveraged optimal transport for cross-modality matching.  
Other methods, such as MULT~\cite{he2024exploring} and DLM~\cite {ye2025dual}, designed a more complex matching scheme by integrating cluster-level matching with instance-level structures to enhance the reliability of cross-modality association.
PCLHD~\cite{shi2024learning} revisited prototype construction in contrastive learning to explore more reliable clustering.
Moreover, ASM~\cite{pang2025augmented} improves the robustness of pseudo labels to color variations by integrating the similarity of augmented images during matching.
For the unpaired setting, MCL~\cite{yao2025unsupervised} generates pseudo cross-modality positive sample pairs through cross-modality feature mapping, constructing a pseudo cross-modality identity space to facilitate effective feature alignment.
Despite their effectiveness, these methods are inherently constrained by the two-stage pipeline, which inevitably introduces modality bias and hinders the modality-invariant learning.

In addition, some methods~\cite{yang2023towards,pang2023cross,yang2024shallow} only perform a single stage of cross-modality learning. 
Specifically, GUR~\cite{yang2023towards} proposed a bottom-up domain learning strategy that performs intra-camera training, inter-camera training, and inter-modality training alternately. 
CHCR~\cite{pang2023cross} designed a cross-modality hierarchical clustering baseline that first refines clusters within each modality before merging them cross-modally based on similarity.
%
SDCL~\cite{yang2024shallow} proposed a shallow-deep collaborative learning framework that initializes with a pre-trained model of single-modality ReID.
Although explicitly abandoning the two-stage pipeline, these methods still suffer from the modality bias issue since they involve single-modality training or clustering.

\subsection{Person ReID with Causal Inference}
Incorporating causal inference \cite{pearl2016causal} into deep learning models, enabling them to learn causal effects, can enhance the performance across various applications.
There has been research exploring the integration of causal inference into person ReID models. For instance, CIFT \cite{li2022counterfactual} utilized counterfactual interventions and causal effect tools to make the graph topology structure more reliable for the VI-ReID graph model. 
Zhang \textit{et al.}\cite{zhang2022learning} approximated causal interventions on domain-specific factors to achieve domain-invariant representation learning for generalizable ReID.  Both AIM \cite{yang2023good} and CCIL \cite{li2023clothes} employed causal intervention models to learn clothing-invariant features for cloth-changing person ReID. These methods cannot be applied in the USL-VI-ReID task since they are designed to mitigate biases caused by domain and clothing rather than modality.

\subsection{Person ReID with Noise Label Learning}
Due to the limited availability of clean annotations in practice, cluster-based unsupervised ReID methods commonly adopt noise label learning mechanisms to refine pseudo-labels and stabilize model training.
%
For example, STDA~\cite{he2023spatial} aggregates spatial-level neighborhood consistency to refine pseudo-labels, while PPLR~\cite{cho2022part} reduces label noise by integrating global and partial predictions with label smoothing.
However, these methods mainly operate on single-modality clustering, leveraging spatial or fine-grained contextual cues, and thus struggle to correct erroneous cross-modality relationships. 
In the USL-VI-ReID task, DPIS~\cite{shi2023dual} and MMM~\cite{shi2024multi} incorporate noisy-label learning by fitting a two-component Gaussian Mixture Model (GMM) to the loss distribution to estimate label confidence, which is then used to penalize noisy samples during optimization. In contrast to such penalization-based strategies, we exploit the estimated confidence to explicitly revise pseudo-labels, thereby mitigating modality bias at the label level rather than merely suppressing its effect.

\section{Proposed Method}

\begin{figure}[t]
    \centering
    \includegraphics[width=1.0\linewidth]{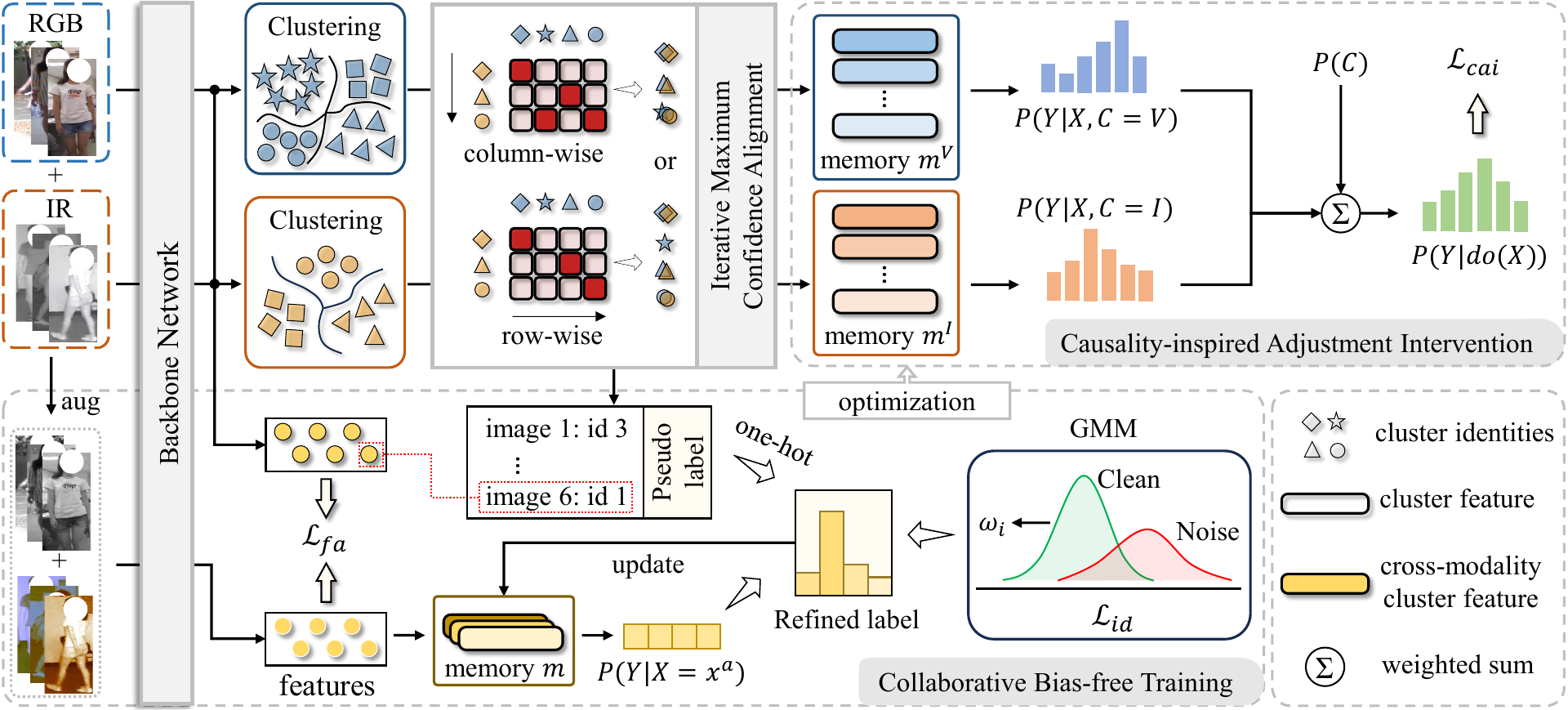}
    \caption{The framework of the proposed DMDL. After obtaining cross-modality pseudo-labels through Iterative Maximum Confidence Alignment, the Causality-inspired Adjustment Intervention module is implemented for causal modeling to construct a low-biased model. Then, the Collaborative Bias-free Training strategy combines label refinement and modality alignment with data augmentation to optimize the model, further eliminating modality bias during training.
    }
    \label{fig:2}
\end{figure}

\subsection{Overview}
The framework of Dual-level Modality Debiasing Learning (DMDL) is shown in Fig.~\ref{fig:2}, incorporating the Causality-inspired Adjustment Intervention (CAI) module and the Collaborative Bias-free Training (CBT) strategy. 
In cross-modality learning, DMDL first iteratively matches clusters across different modalities to obtain cross-modality relationships as a kind of pseudo-label.
Then, CAI employs a backdoor adjustment algorithm to implement causal intervention, which guides the model to capture causal patterns, resulting in a low-biased model.
Furthermore, to avoid misleading optimization caused by biased cues, CBT incorporates label refinement and feature alignment with modality-specific data augmentation to jointly mitigate modality bias across different levels.
This methodology leads to modality-invariant features and a more generalized model.

\subsection{Baseline for Two-stage USL-VI-ReID}
To better illustrate the design of our method and facilitate the organization of experiments, we construct a baseline for two-stage USL-VI-ReID regarding previous works \cite{yang2022augmented,wu2023unsupervised}, which contains a single-modality pre-training stage and a cross-modality learning stage.

The first single-modality learning stage is operated in a clustering-based unsupervised learning manner. Before each training epoch, we first perform clustering on the data from each modality and construct single-modality cluster memories, denoted as $m^c$, by averaging the features of each cluster. $m^c_k$ represents the centroid of cluster $k$ in modality $c\in\{V,I\}$, where $V$ means visible and $I$ is infrared. 
Then, we train the model by contrastive learning on the memory center and corresponding data as follows:

\begin{equation}
\label{eq:id}
\begin{aligned}
    \mathcal{L}^{c}_{id}= - \text{log}\frac{exp(f_x^c \cdot m^{c}_{+}/\sigma)}{\sum\limits_{k} exp(f_x^c \cdot m^{c}_{k}/\sigma)},
\end{aligned}
\end{equation}
where $f_x^c$ is the feature of the image $x$ with the modality $c$, $m_{+}$ is the positive cluster representation, and $\sigma$ is a temperature hyper-parameter. The single-modality model is trained with $ \mathcal{L}^{V}_{id} +  \mathcal{L}^{I}_{id}+{\lambda}_{tri} \cdot  \mathcal{L}_{tri}$, where $\mathcal{L}_{tri}$ is the triplet loss~\cite{hermans2017defense}, and ${\lambda}_{tri}$ controls the weight of $\mathcal{L}_{tri}$ which dynamically changes during training.

In the second stage, we initialize cross-modality learning using the pretrained single-modality model and adopt the clustering-based unsupervised learning pipeline.
To obtain cross-modality pseudo-labels, we propose a simple yet effective \textbf{Iterative Maximum Confidence Alignment (iMCA)} scheme in the baseline to quickly match the $N$ clusters of one modality with the $M$ clusters of the other. 
Let the modality with $N$ clusters be denoted as $C_N$ and the other with $M$ clusters as $C_M$. iMCA first calculates the cosine similarity between cluster centroids to construct an $N \times M$ similarity matrix $S$, where $S_{i,j}$ represents the similarity between the $i$-th cluster of $C_N$ and the $j$-th cluster of $C_M$.
With this, we perform two ways of matching: row-wise and column-wise.
In the row-wise matching, for the $i$-th row of $S$, we find its matched cluster index $u^{row}_i \in [0,M]$ as follows:
\begin{equation}
\label{eq:row_match}
\begin{aligned}
    u^{row}_i = \arg \text{max}_{j} S_{i,j}.
\end{aligned}
\end{equation}
The pseudo-label of the $u^{row}_i$-th cluster in $C_M$ is then assigned to the $i$-th cluster in $C_N$. This operation is applied to all rows ($\forall i \in [0, N]$), effectively propagating labels from $C_M$ to $C_N$.
In the column-wise matching, the same procedure is performed for all columns to propagate labels from $C_N$ to $C_M$.
By alternating between row-wise and column-wise matchings across different epochs, iMCA obtains cross-modality pseudo-labels while preventing the model from being overconfident in a certain matching.
Then, cross-modality cluster memories $m_k$ are established based on unified cross-modality labels for contrastive learning as follows:
\begin{equation}
\label{eq:id_c}
\begin{aligned}
    \mathcal{L}_{id}= - \text{log}\frac{exp(f_x \cdot m_{+}/\sigma)}{\sum\limits_{k} exp(f_x \cdot m_{k}/\sigma)}.
\end{aligned}
\end{equation}
Finally, the unsupervised cross-modality model is trained with $\mathcal{L}_{id}+{\lambda}_{tri} \cdot \mathcal{L}_{tri}$.

\subsection{Modeling with Causal Intervention}
In this section, we first consider USL-VI-ReID from the causal view, analyzing that spurious bias patterns are captured by traditional likelihood-based modeling through a backdoor, whereas causal intervention is not. Based on this analysis, we then illustrate the causal modeling in the proposed CAI module, which constructs a cross-modality model that is insensitive to modality bias.

\subsubsection{USL-VI-ReID from Causal View}
\begin{figure}[t]
    \centering
    \includegraphics[width=0.6\linewidth]{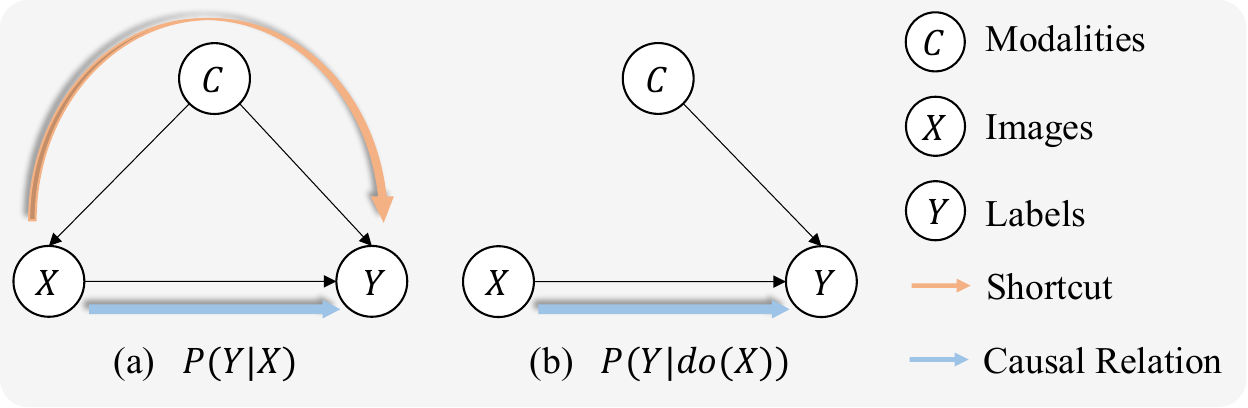}
    \caption{
    (a) The structural causal model in cross-modality learning for USL-VI-ReID. (b) The modified structural causal model after the causal intervention.
    }
    \label{fig:scm}
\end{figure}

To illustrate our motivation for modality debiasing from the causal view, we represent the cross-modality learning process of USL-VI-ReID into the Structural Causal Model (SCM) framework~\cite{pearl2016causal}, as shown in Fig.~\ref{fig:scm} (a). The SCM depicts the relationships among the variables ‘images’ $X$, ‘labels’ $Y$, and ‘modalities’ $C$. 
The arrow $C \to X$ indicates that the modality determines the image pixel values. $X \to Y$ means causal relationships that can recognize human identity from given images. 
Meanwhile, $C \to Y$ reflects the modality bias issue: due to the unsupervised learning pipeline, cross-modality relationships are established based on single-modality clustering and matching that are inherently influenced by modality-specific cues, resulting in biased labels in cross-modality learning.

From this perspective, we can find that modality information influences both the observed images and the inferred labels, inducing a spurious correlation (i.e., a backdoor path) between the input and the prediction, formulated as $X \gets C \to Y$. This backdoor is entangled with the true causal relationship $X \to Y$ and is therefore inevitably captured by the likelihood model, which directly models $P(Y|X)$ without distinguishing causal identity cues from modality-dependent factors. As a result, the learned model tends to exploit modality-induced correlations as shortcuts, resulting in biased predictions and degraded generalization.

To explicitly address this problem, we introduce causal intervention and optimize the interventional distribution $P(Y|do(X))$, as illustrated in Fig.~\ref{fig:scm} (b). The intervention probability $P(Y|do(X=x))$ corresponds to inferring the identity label given an intervened image $X$ fixed to a specific input $x$. The intervention operation $do(\cdot)$ severs the dependency between $X$ and all its potential causes, thereby blocking the path $C \rightarrow X$ and eliminating the backdoor $X \gets C \to Y$. As a result, causal intervention forces the model to rely on identity-related causal patterns rather than modality-specific cues. This provides a principled mechanism for modality debiasing in unsupervised cross-modality learning and motivates our implementation of intervention in CAI to prevent the model from learning modality bias through the backdoor.

\subsubsection{Causality-inspired Adjustment Intervention}
Based on the above analysis, an intervention loss $\mathcal{L}_{cai}$ is constructed by maximizing the intervention probability to eliminate the interference of the modality bias:
\begin{equation}
\label{eq:cai}
\begin{aligned}
    \mathcal{L}_{cai} = \mathbb{E}_{x,y} [-\log P(Y=y|do(X=x))],
\end{aligned}
\end{equation} 
where $x$ denotes an input image, and $y$ represents its associated cross-modality pseudo-label.
To achieve that, CAI implements the computation of $P(Y|do(X)$ by backdoor adjustment~\cite{pearl2016causal} (the detailed derivation is provided in the supplementary material), as follows:
\begin{equation}
\label{eq:backdoor}
\begin{aligned}
    P(Y|do(X))=\sum\limits_{c\in\{V,I\}} P(Y|X,C=c) \cdot P(C=c),
\end{aligned}
\end{equation} 
where $P(C=c)$ means the probabilities of modality $c$, and can be approximated from the training set.
$P(Y|X=x,C=c)$ represents the classification probability of a specific image $x$ inferred by incorporating specific knowledge of modality $c$. Importantly, $c$ is not necessarily the original modality of $x$, which means that the inference needs to combine the image with both visible-specific ($V$) and infrared-specific ($I$) knowledge.
We achieve this by using single-modality memories as follows: 
\begin{equation}
\label{eq:pyfc}
\begin{aligned}
   P(Y=y|X=x,C=c) = \frac{exp(f_x \cdot m^{c}_{y}/\sigma)}{\sum\limits_{k} exp(f_x \cdot m^{c}_{k}/\sigma)},
\end{aligned}
\end{equation}
where $f_x$ is the feature extracted by the backbone model, $y$ represents the cross-modality pseudo-label of the image $x$, and $m^{c}_{y}$ is the cluster centroid of $y$-th cluster of modality $c$.
With these modeled probability parts, we can train the model following Eq.~\eqref{eq:cai}.

We provide further analysis of CAI. Compared to the likelihood model $P(Y|X)$ which can be decomposed as follows:
\begin{equation}
\label{eq:pyx}
\begin{aligned}
    P(Y|X)=\sum\limits_{c} P(Y|X,C=c) \cdot P(C=c|X),
\end{aligned}
\end{equation} 
the backdoor adjustment modifies $P(C = c |X)$ to $P(C = c)$, which can be seen as blocking the correlation between modalities $C$ and images $X$. It eliminates the modality bias during modeling, achieving a low-biased cross-modality model by capturing purely causal relationships.

\subsection{Collaborative Bias-free Training}
Although a low-biased model is obtained through CAI, the biased modality-specific cues existing in labels and features still mislead the model training. To tackle this problem, we propose the CBT strategy to mitigate modality bias at the optimization level. Specifically, considering that modality bias propagates from data into labels and features, CBT integrates label refinement and feature alignment with well-designed data augmentation,  thereby disrupting bias propagation and promoting unbiased feature learning. 

\subsubsection{Data Augmentation in CBT}
\begin{figure}[t]
    \centering
    \includegraphics[width=0.6\linewidth]{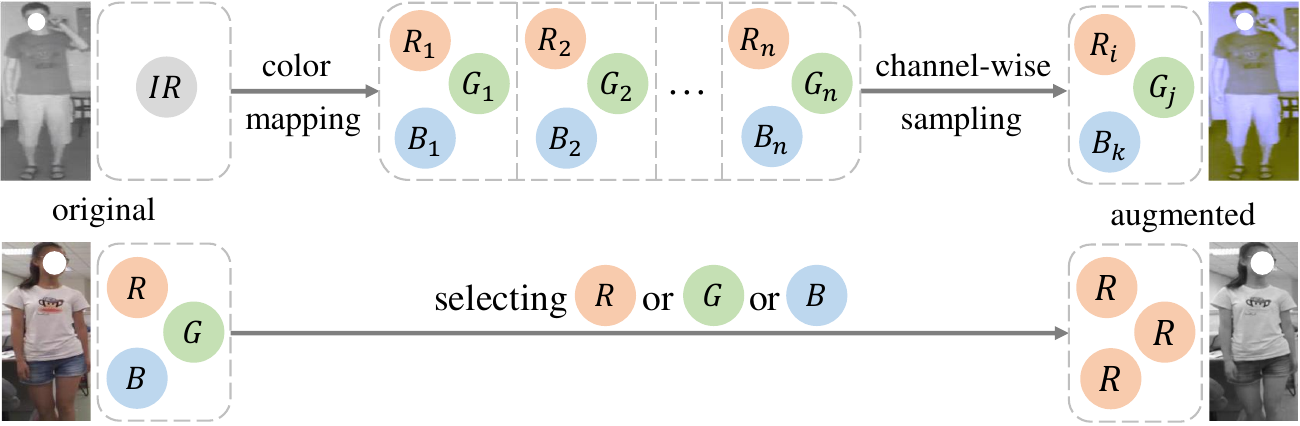}
    \caption{
    Illustration of the modality-specific augmentation. Circles represent channels of images. Subscript represents the sample index of pseudo-color images. For example, $R_2$, $G_2$, and $B_2$ are the red, green, and blue channels from the same pseudo-color image with index 2. The grey circle with ${IR}$ indicates the single channel of the infrared image.
    }
    \label{fig:data}
\end{figure}

CBT first introduces a modality-specific augmentation scheme to destroy modality-related information in images, as shown in Fig.~\ref{fig:data}.
Specifically, for infrared images, we first employ a series of color mapping methods~\cite{bradski2008learning} to transfer each infrared image to multiple pseudo-color images. 
Then, a channel-wise sampling scheme is proposed to increase diversity and introduce randomness to the augmentation by randomly sampling R, G, and B channels of multiple generated pseudo-color images and combining the corresponding sampled channels into a new image. 
For visible images, we employ channel augmentation (CA)~\cite{ye2021channel} through channel multiplexing to generate augmented images, which could derive a series of augmented samples that look like infrared. 

This modality-specific data augmentation enables the image and its corresponding augmentation to share the same identity-discriminative information but differ in modality-related information, mitigating the modality bias at the data level.
With the assistance of such augmentation, CBT implements label refinement and feature alignment to facilitate bias-free learning.

\subsubsection{Label Refinement in CBT}
To refine the noise pseudo-labels, CBT employs label smoothing by exchanging the predictions of images and their augmented images as follows: 
\begin{equation}
\label{eq:refine}
\begin{aligned}
\left\{\begin{array}{l}
\mathbf{\widetilde{y}}_i=w_i \mathbf{y}_i+\left(1-w_i\right) P(Y|X=x_i^a) \\
\mathbf{\widetilde{y}}_i^a=w_i\mathbf{y}_i+\left(1-w_i\right) P(Y|X=x_i),
\end{array}\right.
\end{aligned}
\end{equation}
where $w_i \in [0,1]$ is the refinement weight, representing the reliability of the label $y_i$. The boldface $\mathbf{y}_i$ means the one-hot label vector in which the class index $y_i$ is set to 1 and others are 0. $\mathbf{\widetilde{y}}_i$ and $\mathbf{\widetilde{y}}_i^a$ represent the refined soft labels of image $x_i$ and its augmentation $x_i^a$, respectively. Then, they are used to supervise model training by modifying the $\mathcal{L}_{cai}$ in Eq.~\eqref{eq:cai} as a soft-label classification loss as follows:
\begin{equation}
\begin{aligned}
    \mathcal{L}_{cai} = \mathbb{E}_{i} [&-\sum_k\mathbf{\widetilde{y}}_i[k]\cdot\log P(Y=k|do(X=x_i))\\&-\sum_k\mathbf{\widetilde{y}}^a_i[k]\cdot\log P(Y=k|do(X=x_i^a))],
\end{aligned}
\end{equation} 
where $\mathbf{\widetilde{y}}_i[k]$ means index $k$-th value of $\mathbf{\widetilde{y}}_i$. The computations of $\mathbf{\widetilde{y}}$ and $\mathbf{\widetilde{y}}^a$ depend on $w_i$ and $P(Y|X)$, where the former is the certainty of $y_i$, and the latter is the likelihood function. 

Refer to Eq.~\eqref{eq:refine}, the certainty $w_i$ reflects the reliability of the label for the $i$-th sample. A higher $w_i$ indicates a higher-quality label, allowing $y_i$ to contribute more significantly than $P(Y|X)$ to the final refined label. To quantify the reliability of each label $y_i$ as $w_i$, we follow a common practice in noise label learning~\cite{arazo2019unsupervised} by modeling the distribution of sample losses using a Gaussian Mixture Model (GMM). 
Specifically, we first compute the loss value for each sample via cross-entropy:
\begin{equation}
\label{eq:class}
\begin{aligned}
    \mathcal{L}^i_{id}= - \text{log}\ P(Y=y_i|X=x_i).
\end{aligned}
\end{equation}
The overall loss distribution $\mathcal{L}_{id}$ is then fitted with a two-component GMM, where one component corresponds to low-loss samples (indicating high-quality labels) and the other to high-loss samples (indicating low-quality labels). After training, the GMM estimates the probability that a given loss $\mathcal{L}^i_{id}$ belongs to the low-loss component, which is used as the label certainty $w_i$.

Note that $P(Y|X)$ plays a crucial role in computing both the certainty and the refined labels. We follow Eq.~\eqref{eq:pyx} to implement $P(Y|X)$ but adopt Normalized Weighted Geometric Mean (NWGM)~\cite{xu2015show} for simplification (details are provided in the supplementary material). In one word, $P(Y|X)$ is computed using the modality-shared memory:
\begin{equation}
\begin{aligned}
   P(Y=y|X=x) = \frac{exp(f_x \cdot m_{y}/\sigma)}{\sum\limits_{k} exp(f_x \cdot m_{k}/\sigma)},
\end{aligned}
\end{equation}
where $m_k$ denotes the centroid of the $k$ cluster in the modality-shared memory bank.
It is evident that the quality of the memory bank directly influences the reliability of the predictions. To enhance prediction robustness, we design a dynamic updating scheme that iteratively updates the memory centroid features:
\begin{equation}
\label{eq:memory}
\begin{aligned}
{m}_y \gets \eta_xm_y +  (1-{\eta}_x)f_x,
\end{aligned}
\end{equation}
where $\eta_x$ is an adaptive coefficient determined by:
\begin{equation}
\label{eq:update}
\begin{aligned}
{\eta}_x = {\eta}/{\text{max}(\mathbf{\widetilde{y}}_x[k==y], \eta)}.
\end{aligned}
\end{equation}
Here, $\mathbf{\widetilde{y}}_x[k==y]$ represents the confidence score of sample $x$ being assigned to class $y$, which is obtained from the refined soft label $\mathbf{\widetilde{y}}_x$,
and $\eta$ is a constant threshold set to $0.2$. 
This adaptive coefficient ensures that samples with higher label confidence contribute more substantially to updating the corresponding memory feature $m_y$, while low-confidence samples have limited influence. 

Compared with methods~\cite{shi2023dual,shi2024multi} that only penalize noisy samples based on label certainty, the proposed label refinement directly constructs low-biased cross-modality labels by incorporating modality-specific augmentations and dynamically updating the modality-shared memory. Since an image and its modality-specific augmentation contain different modality-related information, exchanging their predictions for label smoothing effectively reduces label noise induced by modality-specific cues. Furthermore, the dynamic memory updating scheme prevents the memory bank from accumulating noisy representations, thereby ensuring more reliable predictions for refining labels.

\subsubsection{Feature Alignment in CBT}
In addition, a feature alignment loss is introduced to further enhance bias-free feature learning. It is well understood that identity-discriminative information should remain consistent under augmentation. Therefore, for an image and its modality-specific augmentation, the model is expected to extract similar features; otherwise, it suggests that the model is learning modality-specific knowledge.
To this end, we design $\mathcal{L}_{fa}$ following the principles of MMD~\cite{jambigi2021mmd}:
\begin{equation}
\label{eq:pma}
\begin{aligned}
    \mathcal{L}_{fa}= \sum_{c\in\{V,I\}}\lVert \frac{1}{n} \sum^{n}_{i=1} \phi \left( f^c_i\right) - \frac{1}{n} \sum^{n}_{i=1} \phi \left( f^{ca}_i\right) \rVert^{2}_{\mathcal{H}},
\end{aligned}
\end{equation} 
where $f^{ca}_i$ represents the features of the augmented images of modality $c$. $\lVert \cdot \rVert_{\mathcal{H}}$ denotes the distance measured by the Gaussian kernel function $\phi \left( \cdot \right)$, which maps the input to the Reproducing Kernel Hilbert Space (RKHS). This loss constrains the original image and its augmentation representations to be close in the metric space, thereby mitigating the modality-specific cues learned in feature representations. 

\subsection{Total Loss of DMDL}
Following the baseline, the total loss function of our DMDL can be written as: 
\begin{equation}
\label{eq:total}
\begin{aligned}
    \mathcal{L}= \mathcal{L}^{V}_{id} + \mathcal{L}^{I}_{id}+{\lambda}_{cai} \cdot \mathcal{L}_{cai}+{\lambda}_{fa} \cdot \mathcal{L}_{fa}+{\lambda}_{tri} \cdot \mathcal{L}_{tri},
\end{aligned}
\end{equation} 
where ${\lambda}_{cai}$, ${\lambda}_{fa}$ and ${\lambda}_{tri}$ are weights of the corresponding loss term.

\noindent
{\bf Discussion.} In summary, the proposed DMDL framework establishes a unified debiasing pipeline that integrates causal modeling with bias-free optimization. At the modeling level, the CAI module performs causal intervention via backdoor adjustment, encouraging the model to capture causal identity patterns rather than modality-specific shortcuts, thereby constructing a low-biased model. 
Building upon CAI, the CBT further mitigates bias propagation during the optimization process. The modality-specific data augmentation disrupts modality cues at the data level, label refinement corrects biased pseudo-labels at the label level, and feature alignment enforces modality-invariant representations at the feature level. These components collaboratively prevent biased information from being amplified through iterative training.
Importantly, CAI and CBT play complementary roles. CAI suppresses modality bias at the modeling level by reshaping the learning objective, while CBT prevents residual bias from being propagated during optimization. By jointly considering causal intervention and training dynamics, DMDL formulates modality debiasing as an end-to-end learning problem, enabling robust and stable bias suppression throughout the learning pipeline.

\section{Experiments}

\subsection{Datasets and Evaluation Protocol}
\noindent \textbf{Dataset.} 
In this section, we conduct comprehensive experiments to evaluate the proposed method on two widely used datasets, SYSU-MM01 \cite{wu2017rgb} and RegDB \cite{nguyen2017person}, as well as a more recent dataset, LLCM \cite{zhang2023diverse}.

The \textit{SYSU-MM01} dataset with 4 visible cameras and 2 infrared cameras, capturing 395 identities for training and 96 for testing. The test query set comprises 3,803 infrared images, and the gallery set contains 301 visible images. The evaluation protocol provides all-search and indoor-search modes.

The \textit{RegDB} is a dual-camera dataset with 412 identities, each having 10 visible and 10 infrared images. It is split into 206 identities for training and 206 for testing. The evaluation protocol includes two test modes: visible to infrared and infrared to visible.

The \textit{LLCM} is the largest VI-ReID dataset that captures images with 9 cameras. It contains 1,064 identities, of which 713  are used for training and 351 for testing. The evaluation protocol includes two test modes: VIS to IR and IR to VIS.

\vspace{0.1cm}
\noindent \textbf{Evaluation protocol.} All experiments follow the standard evaluation protocol in the VI-ReID benchmark testing. Our model is evaluated using different training/testing splits in ten trials to ensure stable performance. Evaluation metrics include cumulative matching characteristics (CMC), mean average precision (mAP), and mean inverse negative penalty (mINP)~\cite{ye2021deep}.

\subsection{Implementation Details}
We employ ResNet-50 pre-trained on ImageNet as the backbone network and integrate Non-local Attention Blocks~\cite{ye2021deep} and generalized-mean (GeM) pooling~\cite{ye2021deep}.  All input images are resized to 288 × 144, and standard data augmentation techniques, including horizontal flipping, random cropping, and random erasing, are applied.
At the beginning of each epoch, DBSCAN~\cite{ester1996density} clustering is performed independently for each modality to generate pseudo labels. The clustering threshold and the minimum number of images are set to 0.6 and 4 on SYSU-MM01~\cite{wu2017rgb} and LLCM~\cite{zhang2023diverse}, and to 0.3 and 4 on RegDB~\cite{nguyen2017person}, respectively.
During training, 16 pseudo-identities are sampled from each modality, with 16 instances per pseudo-identity (8 original and 8 augmented).
The model is optimized using Adam with an initial learning rate of $3.5 \times 10^{-4}$ and a weight decay of $5 \times 10^{-4}$. The learning rate is decreased by a factor of ten every 20 epochs. The hyperparameter $\sigma$ is set to 0.05. Training proceeds for a total of 100 epochs, with the first 50 epochs dedicated to single-modality learning, followed by 50 epochs of cross-modality training.

\begin{table}[!t]
\centering
\resizebox{\textwidth}{!}{
\begin{tabular}{l|l|cccccc|cccccc}
\hline
\rowcolor[HTML]{EFEFEF} 
\multicolumn{1}{l|}{\cellcolor[HTML]{EFEFEF}}                            & \multicolumn{1}{l|}{\cellcolor[HTML]{EFEFEF}}                        & \multicolumn{6}{c|}{\cellcolor[HTML]{EFEFEF}SYSU-MM01}                                                                                     & \multicolumn{6}{c}{\cellcolor[HTML]{EFEFEF}RegDB}                                                                                   \\  
\rowcolor[HTML]{EFEFEF} 
\multicolumn{1}{l|}{\cellcolor[HTML]{EFEFEF}}                            & \multicolumn{1}{l|}{\cellcolor[HTML]{EFEFEF}}                        & \multicolumn{3}{c|}{\cellcolor[HTML]{EFEFEF}All Search}           & \multicolumn{3}{c|}{\cellcolor[HTML]{EFEFEF}Indoor Search}        & \multicolumn{3}{c|}{\cellcolor[HTML]{EFEFEF}Visible to Infrared}  & \multicolumn{3}{c}{\cellcolor[HTML]{EFEFEF}Infrared to Visible} \\
\rowcolor[HTML]{EFEFEF} 
\multicolumn{1}{l|}{\multirow{-3}{*}{\cellcolor[HTML]{EFEFEF}Methods}}   & \multicolumn{1}{l|}{\multirow{-3}{*}{\cellcolor[HTML]{EFEFEF}Venue}} & r1    & mAP   & \multicolumn{1}{c|}{\cellcolor[HTML]{EFEFEF}mINP} & r1    & mAP   & \multicolumn{1}{c|}{\cellcolor[HTML]{EFEFEF}mINP} & r1    & mAP   & \multicolumn{1}{c|}{\cellcolor[HTML]{EFEFEF}mINP} & r1                 & mAP                 & mINP                \\ \hline
\multicolumn{14}{c}{\textit{Supervised VI-ReID methods}}                                                                                                                                                                                                                                                                                                                                                                                         \\ \hline
\multicolumn{1}{l|}{AGW~\cite{ye2021deep}}         & TPAMI-21   & 47.50  & 47.65  & \multicolumn{1}{c|}{35.30}  & 54.17  & 62.97 & 59.23 & 70.05 & 66.37  & \multicolumn{1}{c|}{50.19}  & 70.49 & 65.90  & 51.24 \\
\multicolumn{1}{l|}{CA~\cite{ye2021channel}}       & ICCV-21    & 69.88  & 66.89  & \multicolumn{1}{c|}{53.61}  & 76.26  & 80.37 & 76.79 & 85.03 & 79.14  & \multicolumn{1}{c|}{65.33}  & 84.75 & 77.82  & 61.56 \\
\multicolumn{1}{l|}{SPOT~\cite{chen2022structure}} & TIP-22     & 65.34  & 62.25  & \multicolumn{1}{c|}{-}      & 69.42  & 74.63 & -     & 80.35 & 72.46  & \multicolumn{1}{c|}{-}      & 79.37 & 72.26  & -     \\ 
\multicolumn{1}{l|}{FMCNet~\cite{zhang2022fmcnet}} & CVPR-22    & 66.34  & 62.51  & \multicolumn{1}{c|}{-}      & 68.15  & 74.09 & -     & 89.12 & 84.43  & \multicolumn{1}{c|}{-}      & 88.38 & 83.86  & -     \\
\multicolumn{1}{l|}{MUN~\cite{yu2023modality}}     & ICCV-23    & 76.24  & 73.81  & \multicolumn{1}{c|}{-}      & 79.42  & 82.06 & -     & 95.19 & 87.15  & \multicolumn{1}{c|}{-}      & 91.86 & 85.01  & -     \\
\multicolumn{1}{l|}{IDKL~\cite{ren2024implicit}}   & CVPR-24    & 81.42  & 79.85  & \multicolumn{1}{c|}{-}      & 87.14  & 89.37 & -     & 94.72 & 90.19  & \multicolumn{1}{c|}{-}      & 94.22 & 90.43  & -     \\ 
\multicolumn{1}{l|}{TSKD~\cite{shi2025two}}        & PR-25    & 76.6  & 73.0  & \multicolumn{1}{c|}{-}      & 82.7  & 85.3 & -     & 91.1 & 81.7  & \multicolumn{1}{c|}{-}      & 89.9 & 80.5  & -     \\ \hline

\multicolumn{14}{c}{\textit{Semi-supervised VI-ReID methods}}                                                                                                                                                                                                                                                                                                                                                                                           \\ \hline
\multicolumn{1}{l|}{OTLA~\cite{wang2022optimal}}    & ECCV-22   & 48.2  & 43.9   & \multicolumn{1}{c|}{-}     & 47.4  & 56.8  & -     & 49.9  & 41.8  & \multicolumn{1}{c|}{-}      & 49.6   & 42.8   & -     \\
\multicolumn{1}{l|}{DPIS~\cite{shi2023dual}}        & ICCV-23   & 58.4   & 55.6  & \multicolumn{1}{c|}{-}     & 63.0  & 70.0  & -     & 62.3  & 53.2  & \multicolumn{1}{c|}{-}      & 61.5   & 52.7   & -     \\ 
\multicolumn{1}{l|}{CGSFL~\cite{zhu2025confidence}} & PR-25     & 59.83  & 53.12 & \multicolumn{1}{c|}{35.79} & 61.50 & 63.83 & 60.66 & 89.36 & 84.17 & \multicolumn{1}{c|}{69.47}  & 89.11  & 81.49  & 66.43  \\ \hline

\multicolumn{14}{c}{\textit{Unsupervised VI-ReID methods}}                                                                                                                                                                                                                                                                                                                                                                                          \\ \hline
\multicolumn{1}{l|}{ADCA~\cite{yang2022augmented}}   & MM-22    & 45.51  & 42.73  & \multicolumn{1}{c|}{28.29}  & 50.60  & 59.11 & 55.17  & 67.20  & 64.05  & \multicolumn{1}{c|}{52.67}  & 68.48  & 63.81  & 49.62  \\
\multicolumn{1}{l|}{DOTLA~\cite{cheng2023unsupervised}} & MM-23 & 50.36  & 47.36  & \multicolumn{1}{c|}{32.40}  & 53.47  & 61.73 & 57.35  & 85.63  & 76.71  & \multicolumn{1}{c|}{61.58}  & 82.91  & 74.97  & 58.60  \\
\multicolumn{1}{l|}{MBCCM~\cite{cheng2023efficient}} & MM-23    & 53.14  & 48.16  & \multicolumn{1}{c|}{32.41}  & 55.21  & 61.98 & 57.13  & 83.79  & 77.87  & \multicolumn{1}{c|}{65.04}  & 82.82  & 76.74  & 61.73      \\
\multicolumn{1}{l|}{PGM~\cite{wu2023unsupervised}}   & CVPR-23  & 57.27  & 51.78  & \multicolumn{1}{c|}{34.96}  & 56.23  & 62.74 & 58.13  & 69.48  & 65.41  & \multicolumn{1}{c|}{52.97}  & 69.85  & 65.17  & -      \\
\multicolumn{1}{l|}{CHCR~\cite{pang2023cross}}       & TCSVT-23 & 59.47  & 59.14  & \multicolumn{1}{c|}{-}      & -      & -     & -      & 69.31  & 64.74  & \multicolumn{1}{c|}{-}      & 69.96  & 65.87  & -      \\
\multicolumn{1}{l|}{GUR$^*$~\cite{yang2023towards}}  & ICCV-23  & 63.51  & 61.63  & \multicolumn{1}{c|}{47.93}  & 71.11  & 76.23 & 72.57  & 73.91  & 70.23  & \multicolumn{1}{c|}{58.88}  & 75.00  & 69.94  & 56.21  \\ 
\multicolumn{1}{l|}{MMM~\cite{shi2024multi}}         & ECCV-24  & 61.6  & 57.9  & \multicolumn{1}{c|}{-}  & 64.4  & 70.4 & -  & 89.7  & 80.5  & \multicolumn{1}{c|}{-}  & 85.8  & 77.0  & -  \\ 
\multicolumn{1}{l|}{PCLHD~\cite{shi2024learning}}    & NIPS-24  & 64.4  & 58.7  & \multicolumn{1}{c|}{-}  & 69.5  & 74.4 & -  & 84.3  & 80.7  & \multicolumn{1}{c|}{-}  & 82.7  & 78.4  & -  \\
\multicolumn{1}{l|}{SDCL$^\dagger$~\cite{yang2024shallow}} & CVPR-24  & 64.49  & 63.24  & \multicolumn{1}{c|}{51.06}  & 71.37  & 76.90 & 73.50  & 86.91  & 78.92  & \multicolumn{1}{c|}{62.83}  & 85.76  & 77.25  & 59.57  \\
\multicolumn{1}{l|}{MULT~\cite{he2024exploring}}     & IJCV-24  & 64.77  & 59.23  & \multicolumn{1}{c|}{43.46}  & 65.34  & 71.46 & 67.83  & 89.95  & 82.09  & \multicolumn{1}{c|}{67.29}  & \textbf{90.78}  & 82.25  & 65.38  \\
\multicolumn{1}{l|}{PCAL~\cite{yang2025progressive}} & TIFS-25  & 57.94  & 52.85  & \multicolumn{1}{c|}{36.90}  & 60.07  & 66.73 & 62.09  & 86.43  & 82.51  & \multicolumn{1}{c|}{72.33}  & 86.21  & 81.23  & 68.71  \\
\multicolumn{1}{l|}{N-ULC~\cite{teng2025relieving}} & AAAI-25  & 61.81  & 58.92  & \multicolumn{1}{c|}{45.01}  & 67.04  & 73.08 & 69.42  & 88.75  & 82.14  & \multicolumn{1}{c|}{68.75}  & 88.17  & 81.11  & 66.05  \\
\multicolumn{1}{l|}{DLM~\cite{ye2025dual}}  & TPAMI-25  & 62.15  & 58.42  & \multicolumn{1}{c|}{43.70}  & 67.31  & 72.86 & 68.89  & 87.55  & 82.83  & \multicolumn{1}{c|}{71.93}  & 86.84  & 81.94  & 68.96  \\
\multicolumn{1}{l|}{RoDE~\cite{10858072}}  & TIFS-25  & 62.88  & 57.91  & \multicolumn{1}{c|}{43.04}  & 64.53  & 70.42 & 66.04  & 88.77  & 78.98  & \multicolumn{1}{c|}{67.99}  & 85.78  & 78.43  & 62.34  \\   
\multicolumn{1}{l|}{SALCR~\cite{cheng2025semantic}}  & IJCV-25  & 64.44  & 60.44  & \multicolumn{1}{c|}{45.19}  & 67.17  & 72.88 & 68.73  & 90.58  & 83.87  & \multicolumn{1}{c|}{70.76}  & 88.69  & 82.66  & 66.89  \\
\multicolumn{1}{l|}{MCL~\cite{yao2025unsupervised}} & ICCV-25  & 62.95  & 62.71  & \multicolumn{1}{c|}{50.63}  &  67.81  &  74.19 &  70.82  &  89.83  &  83.12  & \multicolumn{1}{c|}{ 72.86}  &  88.64  &  82.04  &  69.12  \\
\multicolumn{1}{l|}{ ASM~\cite{pang2025augmented}} &  ICCV-25  &  65.07  &  63.37  & \multicolumn{1}{c|}{ 51.29}  &  71.08  &  76.91 &  73.67  &  88.23  &  79.69  & \multicolumn{1}{c|}{ 63.07}  &  86.85  &  79.20  &  59.96  \\
\hline
\multicolumn{1}{l|}{DMDL}  & Ours      & 65.90  & 61.86  & \multicolumn{1}{c|}{47.53}  & 70.66  & 75.45  & 71.66  
              & \textbf{90.63}  & \textbf{85.33}  & \multicolumn{1}{c|}{\textbf{73.79}}  & 90.30  & \textbf{85.04}  & \textbf{72.00}   \\ 
\multicolumn{1}{l|}{DMDL$^*$}  & Ours  & \textbf{68.04}  & \textbf{65.42}  & \multicolumn{1}{c|}{\textbf{52.01}}  & \textbf{74.81}  & \textbf{79.42}  & \textbf{76.13}  
              & -  & -  & \multicolumn{1}{c|}{-}  & -  & -  & -   \\  \hline
\end{tabular}
}
\caption{
Comparison with the state-of-the-art methods on SYSU-MM01 and RegDB. Rank at r accuracy(\%), mAP (\%) and mINP (\%) are reported. 
$*$ denotes the results of training with extra camera information.
$\dagger$ indicates that the method is initialized with a ReID model pre-trained on additional ReID datasets.
The best results are in \textbf{bold}.
}
\label{tab:comp}
\end{table}

\begin{table}[t]
\centering
\resizebox{0.8\textwidth}{!}{
\begin{tabular}{l|l|cccccc}
\hline
\rowcolor[HTML]{EFEFEF} 
\cellcolor[HTML]{EFEFEF}                          & \multicolumn{1}{l|}{\cellcolor[HTML]{EFEFEF}}                        & \multicolumn{6}{c}{\cellcolor[HTML]{EFEFEF}LLCM}                                                                          \\
\rowcolor[HTML]{EFEFEF} 
\cellcolor[HTML]{EFEFEF}                          & \multicolumn{1}{l|}{\cellcolor[HTML]{EFEFEF}}                        & \multicolumn{3}{c|}{\cellcolor[HTML]{EFEFEF}IR to VIS}            & \multicolumn{3}{c}{\cellcolor[HTML]{EFEFEF}VIS to IR} \\
\rowcolor[HTML]{EFEFEF} 
\multirow{-3}{*}{\cellcolor[HTML]{EFEFEF}Methods} & \multicolumn{1}{l|}{\multirow{-3}{*}{\cellcolor[HTML]{EFEFEF}Venue}} & r1    & mAP   & \multicolumn{1}{c|}{\cellcolor[HTML]{EFEFEF}mINP} & r1               & mAP              & mINP            \\ \hline
\multicolumn{8}{c}{\textit{ Supervised VI-ReID methods}}                                                                                                             \\ \hline
\multicolumn{1}{l|}{AGW \cite{ye2021deep}}        & TPAMI-21   & 43.6  & 51.8  & \multicolumn{1}{c|}{-}    & 51.5  & 55.3  & -      \\
\multicolumn{1}{l|}{LbA \cite{park2021learning}}  & ICCV-21    & 43.8  & 53.1  & \multicolumn{1}{c|}{-}    & 50.8  & 55.6  & -      \\
\multicolumn{1}{l|}{CA \cite{ye2021channel}}      & ICCV-21    & 48.8  & 56.6  & \multicolumn{1}{c|}{-}    & 56.5  & 59.8  & -      \\
\multicolumn{1}{l|}{DART \cite{yang2022learning}} & CVPR-22    & 52.2  & 59.8  & \multicolumn{1}{c|}{-}    & 60.4  & 63.2  & -      \\
\multicolumn{1}{l|}{DEEN \cite{zhang2023diverse}} & CVPR-23    & 54.9  & 62.9  & \multicolumn{1}{c|}{-}    & 62.5  & 65.8  & -      \\ 
\multicolumn{1}{l|}{CM\textsuperscript{2}GT~\cite{feng2025learning}}  & PR-25    & 52.1  & 58.3  & \multicolumn{1}{c|}{-}      & 65.9  & 50.3 & -    \\ \hline
\multicolumn{8}{c}{\textit{ Unsupervised VI-ReID methods}}                                                                                                                 \\ \hline
\multicolumn{1}{l|}{ADCA~\cite{yang2022augmented}}   & MM-22    & 23.57  & 28.25  & \multicolumn{1}{c|}{-}    & 16.16  & 21.48  & -      \\
\multicolumn{1}{l|}{DOTLA~\cite{cheng2023unsupervised}} & MM-23 & 27.14  & 26.26  & \multicolumn{1}{c|}{-}    & 23.52  & 27.48  & -      \\
\multicolumn{1}{l|}{GUR~\cite{yang2023towards}}  & ICCV-23  & 31.47  & 34.77  & \multicolumn{1}{c|}{-}    & 29.68  & 33.38  & -      \\
\multicolumn{1}{l|}{IMSL~\cite{pang2024inter}}       & TCSVT-24 & 22.74  & 19.38  & \multicolumn{1}{c|}{-}    & 17.26  & 24.38  & -      \\
\multicolumn{1}{l|}{RoDE \cite{10858072}}         & TIFS-25    & 32.73  & 36.64  & \multicolumn{1}{c|}{-}    & 35.13  & 37.44  & -      \\ \hline
\multicolumn{1}{l|}{DMDL}                         & Ours       & \textbf{45.25} & \textbf{50.87} & \multicolumn{1}{c|}{\textbf{47.14}}       & \textbf{51.84}  & \textbf{55.35}  & \textbf{49.85}   \\ \hline   
\end{tabular}
}
\caption{
Comparison with the state-of-the-art methods on the LLCM dataset. Rank at r accuracy(\%), mAP (\%) and mINP (\%) are reported.
The best results are in \textbf{bold}.
}
\label{tab:llcm}
\end{table}

\subsection{Comparison with State-of-the-art Methods}
To validate the effectiveness of our DMDL, we compare it with state-of-the-art methods under three relevant settings: supervised VI-ReID, semi-supervised VI-ReID, and unsupervised VI-ReID. The experimental results for the SYSU-MM01 and RegDB datasets are shown in Table~\ref{tab:comp}, and the experimental results for the LLCM dataset are presented in Table~\ref{tab:llcm}.

\vspace{0.1cm}
\noindent \textbf{Comparison with supervised VI-ReID Methods.} 
Encouragingly, our DMDL achieves competitive performance compared to the supervised method FMCNet~\cite{zhang2022fmcnet} on the SYSU-MM01 and RegDB datasets, and even surpasses several supervised methods, including AGW~\cite{ye2021deep} and SPOT~\cite{chen2022structure}.
Moreover, on the challenging LLCM dataset, our DMDL still demonstrates impressive performance, outperforming several supervised methods (e.g., AGW~\cite{ye2021deep} and LbA~\cite{park2021learning}).
However, due to the absence of annotated cross-modality correspondences, unsupervised methods still have significant room for improvement compared to the state-of-the-art supervised VI-ReID methods.

\vspace{0.1cm}
\noindent \textbf{Comparison with semi-supervised VI-ReID Methods.} 
Semi-supervised VI-ReID methods are trained on datasets with partial annotations. Remarkably, our DMDL achieves outstanding performance without relying on any annotations, surpassing all semi-supervised counterparts, as reported in Table~\ref{tab:comp}. These results highlight the potential of USL-VI-ReID, which eliminates the need for annotations and offers greater practicality.

%
\vspace{0.1cm}
\noindent \textbf{Comparison with unsupervised VI-ReID Methods.} 
The results in Table~\ref{tab:comp} demonstrate that our method achieves superior performance under the unsupervised VI-ReID setting. Specifically, DMDL attains 65.42\% mAP on SYSU-MM01 (all-search) and 85.33\% mAP on RegDB (visible-to-infrared), outperforming the method SALCR~\cite{cheng2025semantic} by 4.98\% and 1.46\% mAP on the respective datasets.
Notably, even without utilizing camera information, our approach achieves rank-1 accuracies of 65.90\% on SYSU-MM01 (all-search) and 90.63\% on RegDB (visible-to-infrared), respectively. These results surpass those of existing methods, including the recent state-of-the-art approaches MCL~\cite{yao2025unsupervised} and ASM~\cite{pang2025augmented}, demonstrating the strong effectiveness of our method.
Meanwhile, on the more challenging LLCM dataset, our method showcases remarkable performance, surpassing the state-of-the-art RoDE~\cite{10858072} by 14.23\% mAP and 12.52\% rank-1 accuracy in the IR to VIS setting, as reported in Table~\ref{tab:llcm}.

These results suggest that while existing methods have made notable progress on the USL-VI-ReID task, they still suffer from modality bias, leading to the extraction of modality-dependent features. In contrast, our DMDL effectively learns modality-invariant representations by addressing modality bias through causal modeling and bias-free training optimization, achieving more robust performance.

\begin{table}[!t]
\resizebox{\textwidth}{!}{
\begin{tabular}{c|ccccc|cccccccccc}
\hline
\rowcolor[HTML]{EFEFEF} 
                                                & \multicolumn{5}{c|}{\cellcolor[HTML]{EFEFEF}Components}                                                                                               & \multicolumn{10}{c}{\cellcolor[HTML]{EFEFEF}SYSU-MM01}                                                                              \\ \hline
\rowcolor[HTML]{EFEFEF} 
\cellcolor[HTML]{EFEFEF}                        & \cellcolor[HTML]{EFEFEF}                           & \cellcolor[HTML]{EFEFEF}                      & \multicolumn{3}{c|}{\cellcolor[HTML]{EFEFEF}CBT} & \multicolumn{5}{c|}{\cellcolor[HTML]{EFEFEF}All Search}                 & \multicolumn{5}{c}{\cellcolor[HTML]{EFEFEF}Indoor Search} \\
\rowcolor[HTML]{EFEFEF} 
\multirow{-2}{*}{\cellcolor[HTML]{EFEFEF}Index} & \multirow{-2}{*}{\cellcolor[HTML]{EFEFEF}Baseline} & \multirow{-2}{*}{\cellcolor[HTML]{EFEFEF}CAI} & data          & label          & feature         & r1 & r5 & r10 & mAP & \multicolumn{1}{c|}{\cellcolor[HTML]{EFEFEF}mINP} & r1        & r5       & r10       & mAP       & mINP       \\ \hline  \hline
1     & \checkmark  &            &            &            &            
& 56.26  & 84.12  & 92.19  & 54.60  & \multicolumn{1}{c|}{40.75}                         & 63.98  & 87.77  & 94.06  & 69.68  & 65.30    \\ 
2     & \checkmark  & \checkmark &            &            &            
& 59.61  & 85.34  & 92.94  & 57.85  & \multicolumn{1}{c|}{44.31}                         & 67.07  & 89.67  & 95.02  & 72.94  & 69.00    \\
3     & \checkmark  & \checkmark & \checkmark &            &            
& 62.17  & 87.01  & 93.82  & 58.76  & \multicolumn{1}{c|}{45.59}                         & 68.07  & 89.92  & 94.51  & 73.08  & 69.25      \\
4     & \checkmark  & \checkmark & \checkmark & \checkmark &            
& 63.11  & 87.67  & 94.66  & 59.61  & \multicolumn{1}{c|}{45.34}                         & 68.34  & 90.22  & 95.29  & 73.06  & 69.09    \\
5     & \checkmark  & \checkmark & \checkmark &            & \checkmark           
& 64.09  & 88.07  & 94.65  & 61.19  & \multicolumn{1}{c|}{47.41}                         & 68.56  & 90.36  & 95.27  & 73.91  & 70.08     \\
6     & \checkmark  &            & \checkmark & \checkmark & \checkmark           
& 63.39  & 87.88  & 93.85  & 59.78  & \multicolumn{1}{c|}{45.08}                         & 67.26  & 90.31  & 95.52  & 72.77  & 68.44     \\
\rowcolor[HTML]{ECF4FF} 
7     & \checkmark  & \checkmark & \checkmark & \checkmark & \checkmark & \textbf{65.90} & \textbf{88.56} & \textbf{94.71} & \textbf{61.86} & \multicolumn{1}{c|}{\cellcolor[HTML]{ECF4FF}\textbf{47.53}} & \textbf{70.66}  & \textbf{91.18} & \textbf{95.78}    & \textbf{75.45}    & \textbf{71.66}   \\ \hline
\end{tabular}
}
\caption{
Ablation studies on the SYSU-MM01. Rank at r accuracy (\%), mAP (\%) and mINP (\%) are reported.
}
\label{tab:abla}
\end{table}

\subsection{Ablation Study}
To evaluate the contribution of each component in DMDL, we conduct ablation experiments on the SYSU-MM01 dataset, as summarized in Table~\ref{tab:abla}.
Note that channel augmentation (CA)~\cite{ye2021channel} for visible images is incorporated into the \emph{baseline} to ensure a fair assessment of our designed components.

\vspace{0.1cm}
\noindent \textbf{Effectiveness of CAI.} 
When CAI is applied by replacing traditional likelihood-based modeling in \emph{baseline} with causal modeling, the rank-1 accuracy and mAP of all-search increase by 3.35\% and 3.25\%, respectively (Index 1 vs. 2).
This demonstrates that CAI effectively constructs a low-biased model by explicitly modeling causal relationships between images and labels, enhancing the model’s robustness to modality variation.

\vspace{0.1cm}
\noindent \textbf{Discussion of removing CAI.} 
We conduct the experiments of removing CAI and only using CBT with baseline (Index 6). Compared with the full DMDL (CAI+CBT), using CBT alone degrades the rank-1 accuracy by approximately 2–3\%, suggesting that  CAI and CBT are most effective when used jointly. CAI constructs a low-biased model, while CBT suppresses the injection of biased cues into the model during optimization. They act on the complementary modeling and optimization stages, and reinforce each other in modality debiasing.

\vspace{0.1cm}
\noindent \textbf{Effectiveness of CBT.} 
Compared with the results in Index 2, the experiments in Index 7 show that the proposed optimization strategy, CBT, achieves a 6.29\% improvement in rank-1 accuracy (all-search), confirming its effectiveness in bias-free feature learning.
Specifically, CBT comprises three components: data augmentation, label refinement, and feature alignment.
When integrated sequentially, these components yield consistent performance gains (Index 3–7). This trend indicates that each component mitigates modality bias at different levels, contributes to learn modality-invariant representations, and produces a mutually reinforcing effect.
Below, we present a detailed analysis of these components in CBT.

\begin{figure}[t]
    \centering
    \includegraphics[width=1.0\linewidth]{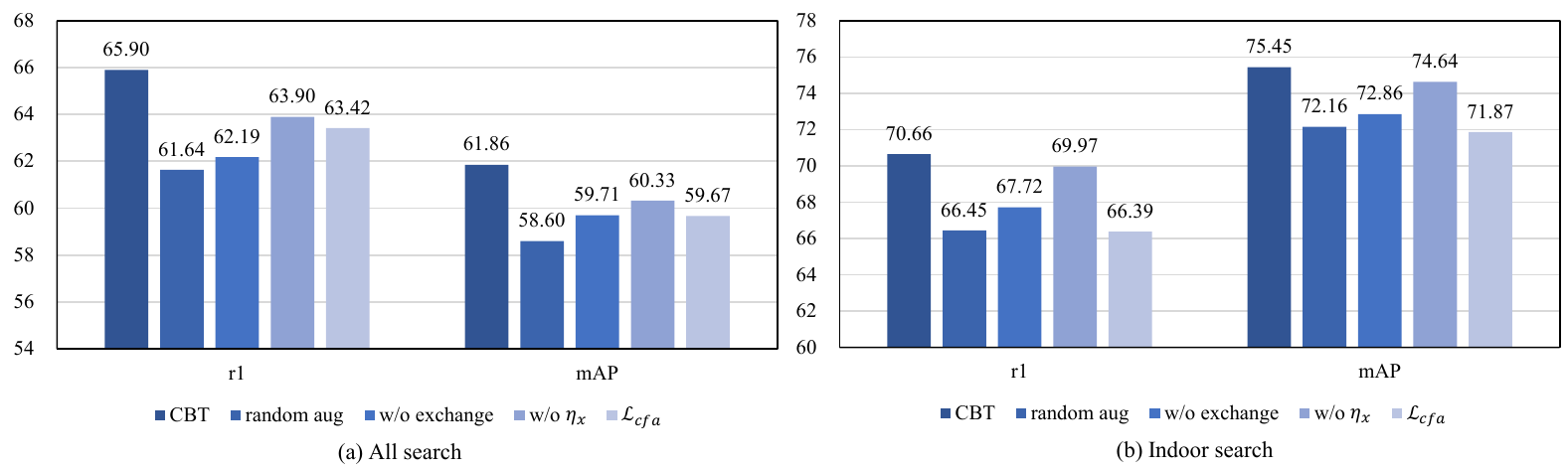}
    \caption{
   Detailed analysis of CBT on the SYSU-MM01 dataset under (a) all-search and (b) indoor-search modes. Rank-1 accuracy (\%) and mAP (\%) are reported. }
    \label{fig:cbt}
\end{figure}

\vspace{0.1cm}
\noindent \textbf{Effectiveness of data augmentation in CBT.} 
To verify the effectiveness of modality-specific augmentation, we replace it with standard random augmentation schemes (e.g., random cropping) within the CBT strategy. As shown in the second column ``random aug'' of Fig.~\ref{fig:cbt}, this replacement results in a drop of about 4\% in rank-1 accuracy. This drop indicates that modality-specific augmentation plays a crucial role in CBT, as it explicitly disrupts modality-specific information in the images and thus guides the model to learn modality-shared representations.
Moreover, when only the modality-specific augmentation is applied, the model still achieves an overall performance improvement (Index 3 vs. 4), further demonstrating that this augmentation effectively mitigates modality bias at the data level.

\vspace{0.1cm}
\noindent \textbf{Effectiveness of label refinement in CBT.} 
Based on the data augmentation, the label refinement scheme introduces two key designs:
(1) exchanging predictions between an image and its augmentation, and
(2) dynamically updating the memory with adaptive $\eta_x$.
To evaluate the effectiveness of these designs, we conduct ablation experiments under the “w/o exchange” and “w/o $\eta_x$” settings, the results of which are shown in the third and fourth columns of Fig.~\ref{fig:cbt}.
Specifically, the “w/o exchange” variant refines the pseudo-label using its own prediction rather than that of its augmentation, 
while the “w/o $\eta_x$” variant fixes $\eta=0.05$ for memory updating.
Both variants lead to performance degradation, confirming that:
(1) exchanging predictions between images and their modality-specific augmentations mitigates modality bias in refined labels, as the augmentation perturbs modality-specific cues, yielding less biased predictions for refining labels, and (2) dynamic updating enhances memory reliability, enabling stable predictions.
By integrating these two designs, CBT effectively mitigates modality bias at the label level, resulting in a substantial performance improvement (Index 3 vs. 4).

\vspace{0.1cm}
\noindent \textbf{Effectiveness of feature alignment in CBT.} 
We observe a notable performance gain after incorporating the feature alignment loss $\mathcal{L}_{fa}$ into CBT (Index 3 vs. 5), indicating that  $\mathcal{L}_{fa}$ effectively alleviates modality bias in feature representations by aligning images and their augmentation.
To further validate the effectiveness of our design, we replace $\mathcal{L}_{fa}$ with an MMD-based loss $\mathcal{L}_{cfa}$ commonly adopted in conventional VI-ReID methods, which enforces direct alignment between visible and infrared feature distributions to reduce modality gap:
\begin{equation}
\begin{aligned}
    \mathcal{L}_{cfa}= \lVert \frac{1}{n} \sum^{n}_{i=1} \phi \left( f^V_i\right) - \frac{1}{n} \sum^{n}_{i=1} \phi \left( f^{I}_i\right) \rVert^{2}_{\mathcal{H}}.
\end{aligned}
\end{equation}
As shown in the fifth column of Fig.~\ref{fig:cbt}, $\mathcal{L}_{cfa}$ results in degraded performance, since inconsistent cross-modality pseudo-labels lead to identity misalignment and weaken feature discriminability. In contrast, $\mathcal{L}_{fa}$ leverages the natural correspondence between original and augmented images, aligning modalities in a label-consistent and feature-discriminative manner.
Furthermore, combining augmentation-based label refinement and feature alignment achieves the best result (Index 7), indicating that CBT effectively promotes bias-free feature learning by interrupting the propagation of modality bias across data, labels, and features.

\subsection{Further Analysis}

\begin{figure}[!t]
    \centering
    \includegraphics[width=0.75\linewidth]{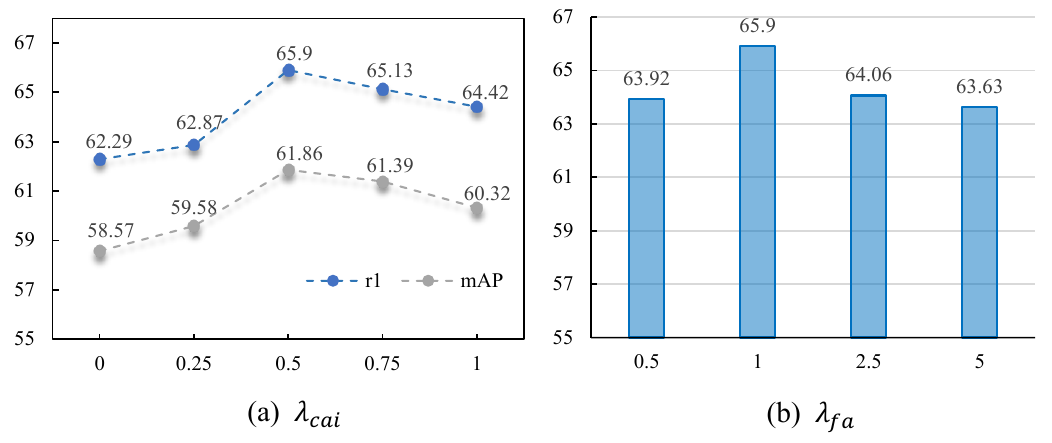}
    \caption{
   Parameter analysis of ${\lambda}_{cai}$ and ${\lambda}_{fa}$ on the SYSU-MM01 dataset (all-search). }
    \label{fig:para}
\end{figure}

\begin{table}[!t]
\centering
\resizebox{0.6\textwidth}{!}{
\begin{tabular}{l|ccc|ccc}
\hline
\rowcolor[HTML]{EFEFEF} 
SYSU-MM01 & \multicolumn{3}{c|}{\cellcolor[HTML]{EFEFEF}All Search} & \multicolumn{3}{c}{\cellcolor[HTML]{EFEFEF}Indoor Search} \\ \hline 
\rowcolor[HTML]{EFEFEF} 
methods    & r1                & mAP              & mINP             & r1                & mAP               & mINP              \\ \hline \hline
OT~\cite{cheng2023unsupervised}        & 60.24             & 55.13            & 38.24            & 64.13             & 68.64             & 62.43             \\
BGM~\cite{wu2023unsupervised}        & 63.74             & 58.87            & 42.88            & 68.57             & 73.06            & 68.48             \\
iMCA(ours)      & \textbf{65.90}             & \textbf{61.86}            & \textbf{47.53}            & \textbf{70.66}             & \textbf{75.45}             & \textbf{71.66}             \\ \hline
\end{tabular}
}
\caption{
 The comparison of different matching strategies on SYSU-MM01. Rank1 accuracy (\%), mAP (\%) and mINP (\%) are reported.
}
\label{tab:match}
\end{table}

\vspace{0.1cm}
\noindent \textbf{Parameter Analysis.} 
The proposed DMDL introduces two key parameters, ${\lambda}_{cai}$ and ${\lambda}_{fa}$ in Eq.~\ref{eq:total}, which serve as weighting factors to balance $\mathcal{L}_{cai}$ and $\mathcal{L}_{fa}$ during training.
Fig.~\ref{fig:para} (a) illustrates the impact of varying ${\lambda}_{cai}$ on rank-1 and mAP accuracy on the SYSU-MM01 dataset (all-search). When ${\lambda}_{cai}=0$, the CAI is disabled, resulting in poor performance, which confirms its effectiveness. The best performance is observed at ${\lambda}_{cai}=0.5$, and this value is therefore adopted in our experiments.
Fig.~\ref{fig:para} (b) shows the rank-1 accuracy results for different ${\lambda}_{fa}$ values, and the model achieves the highest accuracy at ${\lambda}_{fa}=1$, so we empirically set it to 1.
For completeness, the sensitivity analysis of ${\lambda}_{tri}$ used in \emph{baseline} is provided in the supplementary material, as it is not our main contribution.

\begin{figure}[!t]
    \centering
    \includegraphics[width=0.7\linewidth]{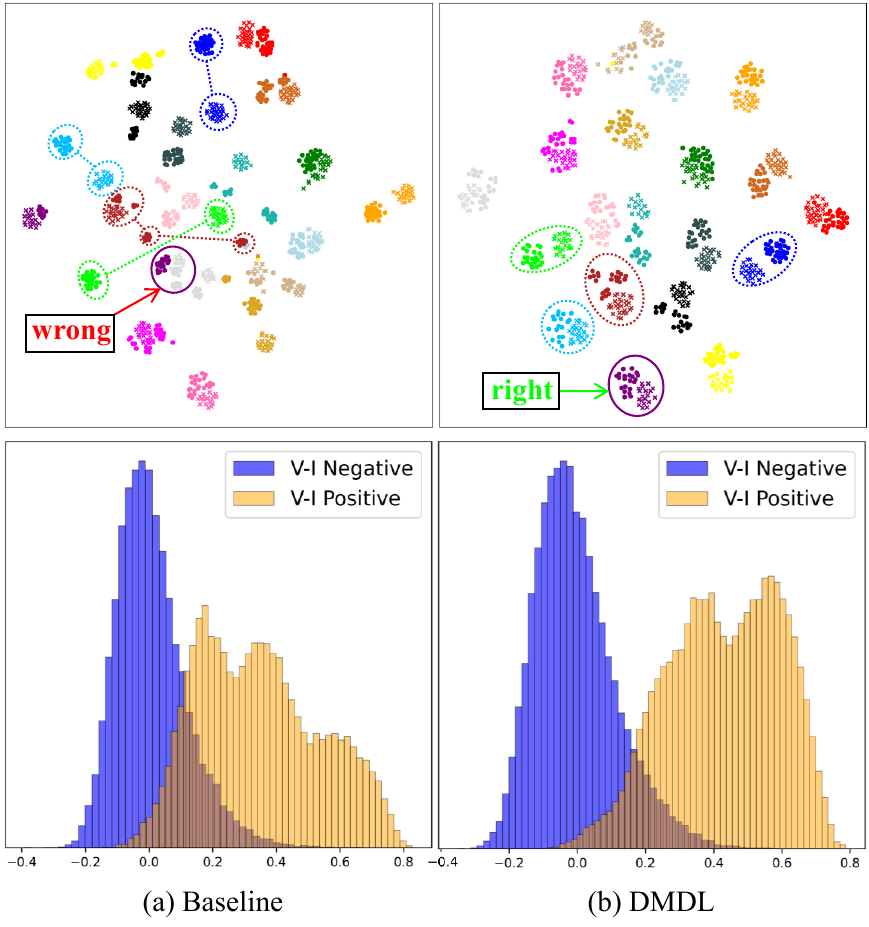}
    \caption{
   The t-SNE (first row) and similarity distribution (second row) visualization of 20 randomly selected identities on the SYSU-MM01 dataset. In t-SNE visualization, the circle and the cross represent the visible and infrared modalities, respectively. }
    \label{fig:4}
\end{figure}

\vspace{0.1cm}
\noindent \textbf{Visualization Analysis.} 
Fig.~\ref{fig:4} presents the t-SNE~\cite{van2008visualizing} plots and the cosine similarity distribution of positive and negative cross-modality pairs for randomly selected identities. 
Compared with the baseline, DMDL exhibits a more compact alignment between visible and infrared samples, together with a larger separation between positive and negative cross-modality pairs. In addition, the mismatched samples highlighted by the purple circle are correctly aligned after applying DMDL. 
Taken together, these visualizations demonstrate that DMDL effectively narrows the modality gap and improves the robustness of the learned representation against modality bias.

\begin{figure}[t]
    \centering
    \includegraphics[width=0.7\linewidth]{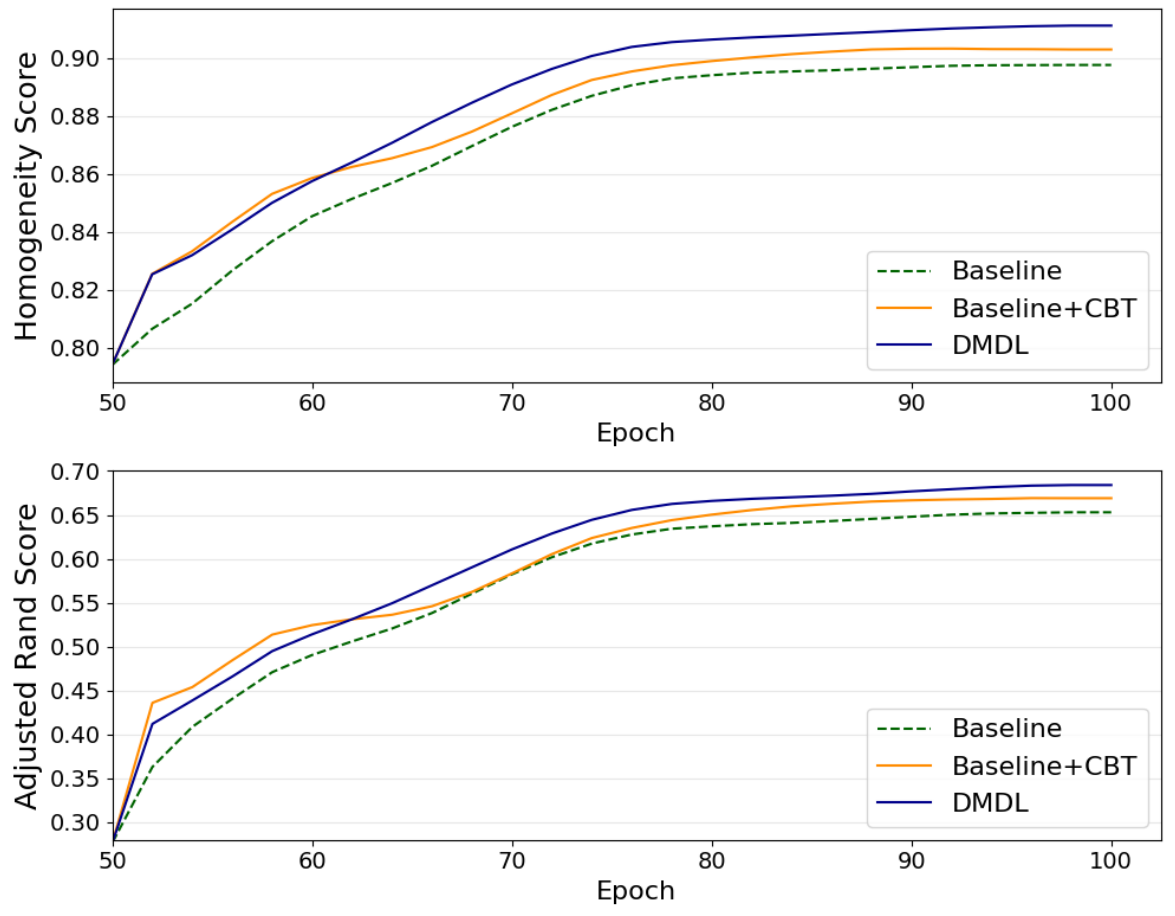}
    \caption{
   Cross-modality pseudo-label quality analysis over different epochs on the SYSU-MM01 dataset. }
    \label{fig:match}
\end{figure}

\vspace{0.1cm}
\noindent \textbf{Cross-modality Pseudo-label Quality Analysis.} 
We assess the quality of cross-modality pseudo-labels generated at different training epochs on the SYSU-MM01 dataset in Fig.~\ref{fig:match}, using two standard metrics from~\cite{scikit-learn}: Homogeneity Score and Adjusted Rand Score, where higher scores indicate better label quality. Notably, incorporating CBT and CAI on top of the baseline consistently improves the quality of cross-modality labels across training. This demonstrates that CBT and CAI effectively mitigate modality bias in the learned representations, thereby facilitating more accurate cross-modality matching.

\vspace{0.1cm}
\noindent \textbf{Effectiveness of iMCA.}
To evaluate the robustness of our matching strategy, iMCA, we compare it with two commonly adopted matching strategies, bipartite graph matching (BGM) and optimal transport (OT), under identical experimental settings, as reported in Table~\ref{tab:match}. Although both BGM and OT are capable of establishing cross-modality correspondences, their performance is consistently inferior to that of iMCA. Specifically, OT assigns samples to cross-modality clusters under an implicit uniform assignment assumption, while BGM enforces a strict global one-to-one cluster matching. These strong assumptions make both methods more prone to erroneous assignments, especially in the presence of noisy or ambiguous cross-modality similarities. In contrast, our iMCA performs conservative cross-modality alignment through a natural maximum-confidence matching mechanism without any assumptions, leading to more stable and reliable correspondences.

\begin{figure*}[t]
    \centering
    \includegraphics[width=1.0\linewidth]{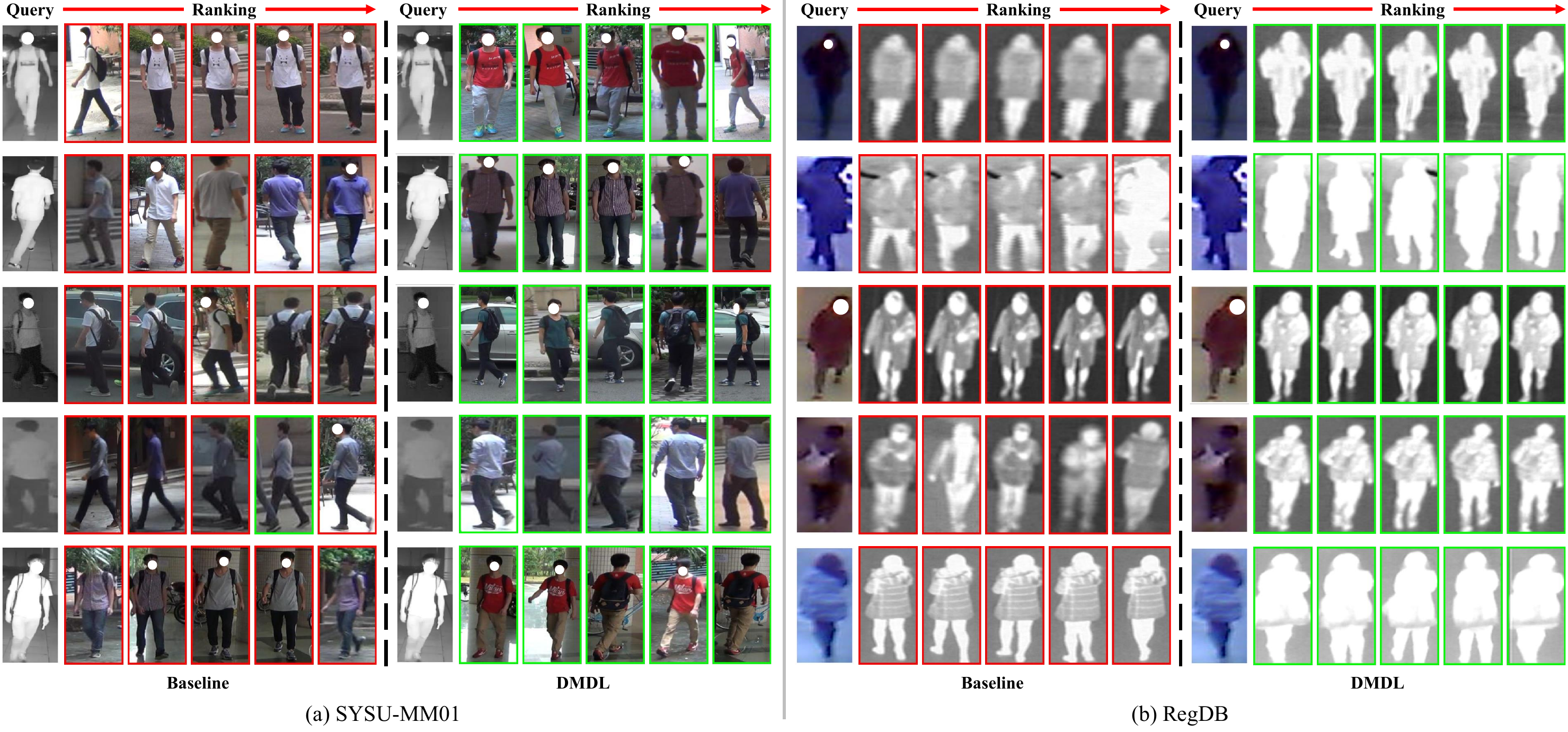}
    \caption{Visualization of the retrieval results obtained by the baseline and our DMDL on the SYSU-MM01 and RegDB datasets. The green boxes represent correct retrieval results, and the red boxes represent incorrect retrieval results. 
    }
    \label{fig:rank}
\end{figure*}

\begin{figure*}[t]
    \centering
    \includegraphics[width=1.0\linewidth]{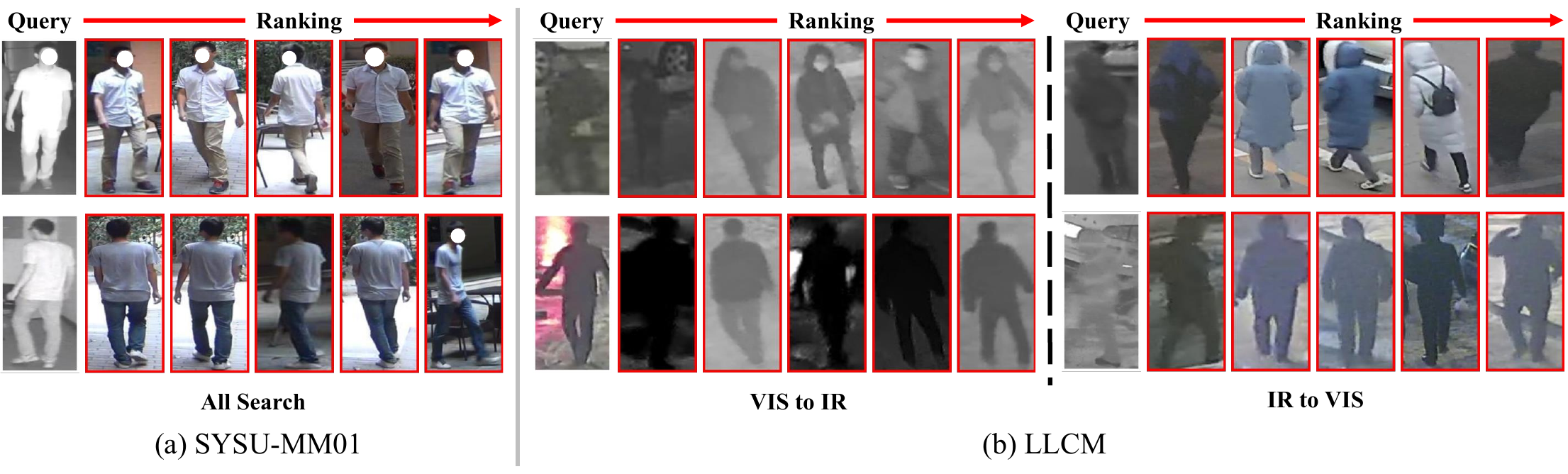}
   \caption{Visualization of representative failure examples on the SYSU-MM01 and LLCM datasets.}
    \label{fig:neg_rank}
\end{figure*}

\vspace{0.1cm}
\noindent \textbf{Retrieval Results.}
We qualitatively compare our DMDL with the baseline by visualizing the retrieval results of several query images on SYSU-MM01 and RegDB, as illustrated in Fig.~\ref{fig:rank}. For each query, the retrieved samples highlighted with green boxes indicate correct matches, while those marked in red correspond to incorrect matches. 
Overall, the proposed method exhibits higher robustness to modality-specific interference (e.g., color cues), whereas the baseline tends to prioritize color similarity when retrieving results (see the first row in Fig.~\ref{fig:rank}). This confirms that our method achieves stronger cross-modality retrieval capability than the baseline, yielding consistent improvements.

\vspace{0.1cm}
\noindent \textbf{Failure-case Analysis.}
From the challenging examples shown in Fig.~\ref{fig:neg_rank}, we observe that performance degradation mainly occurs under extremely difficult conditions, such as severe occlusion, low resolution, and heavy background clutter. Similar failure cases are also observed in the USL-VI-ReID method RoDE \cite{10858072}, indicating that such performance degradation stems from the inherent challenges of unsupervised VI-ReID. When identity-discriminative cues in the query modality are weak or partially missing, models struggle to extract sufficiently informative representations, which consequently leads to degraded matching accuracy. In future work, incorporating richer causal structures, such as explicitly modeling environment-related factors, may help alleviate these limitations and further improve robustness under extreme conditions.

From the challenging examples shown in Fig.~\ref{fig:neg_rank}, we observe that performance degradation mainly occurs under extremely difficult conditions, such as severe occlusion, low resolution, and heavy background clutter. These cases are largely attributed to the inherent challenges of both supervised and unsupervised VI-ReID. When identity-discriminative cues in the query modality are weak or partially missing, the model struggles to extract sufficiently informative representations, which consequently degrades matching accuracy. In future work, incorporating richer causal structures, such as explicitly modeling environment-related factors, may help alleviate these limitations and further improve robustness under extreme conditions.
\section{Conclusion}
In this paper, we investigate the modality bias issue in unsupervised VI-ReID and propose a novel Dual-level Modality Debiasing Learning (DMDL) framework to tackle this issue from both model and optimization perspectives, incorporating a Causality-inspired Adjustment Intervention (CAI) module and a Collaborative Bias-free Training (CBT) strategy. 
CAI models causal relationships between images and pseudo-labels to capture stable, modality-independent patterns, thereby constructing a low-biased model. 
Meanwhile, CBT performs label refinement and feature alignment with modality-specific data augmentation, jointly preventing the propagation of modality bias and thus achieving bias-free optimization.
Finally, with the above designs, DMDL effectively achieves modality-invariant feature learning. Extensive experiments on benchmark datasets validate the superior performance of our method. 

\noindent\textbf{Limitations.}
Despite the proposed framework demonstrating strong performance, several limitations remain. First, the adopted causal graph focuses on dominant modality bias and remains relatively simplified. Incorporating richer causal structures, such as camera-specific or environment-related factors, may further improve the interpretability and effectiveness of causal intervention in cross-modality re-identification.
Second, although iMCA provides a robust initialization for cross-modality alignment, the framework still relies on the quality of early-stage pseudo-labels, which may affect convergence in extremely challenging scenarios. More adaptive or curriculum-based alignment strategies could be explored to further enhance robustness.
Finally, the pseudo-color augmentation is designed to disrupt superficial modality-specific cues rather than to generate realistic visible images under complex real-world conditions. Its effectiveness may therefore be limited in scenarios dominated by environmental factors, such as low-light or cluttered backgrounds. Future work could investigate more advanced generation or simulation-based strategies.

\bibliographystyle{elsarticle-num} 
\bibliography{cas-refs}






\end{document}